\documentclass[journal]{IEEEtran}

\usepackage{amsmath,amssymb,amsfonts}
\usepackage{algorithmic}
\usepackage{graphicx}
\usepackage{textcomp}
\usepackage{multirow}
\usepackage[colorlinks=true, linkcolor=blue]{hyperref} 
\usepackage{booktabs} 
\usepackage{adjustbox} 

\usepackage{float}

\usepackage{adjustbox}
\usepackage{tabularx}
\ifCLASSINFOpdf
\else
\fi
\begin{document}
%
\title{LGPS: A Lightweight GAN-Based Approach for Polyp Segmentation in Colonoscopy Images}
%
%
%

\author{Fiseha B. Tesema, \IEEEmembership{Member, IEEE}, Alejandro Guerra Manzanares ,Tianxiang Cui,  Qian Zhang, Moses Solomon, Sean He
\thanks{This work was supported by RKE Internal Seed Funding for a Small Research and Knowledge Exchange Project [105231000022] and Nottingham Ningbo China Beacons of Excellence Research and Innovation Institute[I01240100007], University of Nottingham Ningbo China(UNNC), Ningbo, China. (Corresponding author: Fiseha B. Tesema)}
\thanks{Fiseha B. Tesema is with the School of Computer Science and the Nottingham Ningbo China Beacons of Excellence Research and Innovation Institute, UNNC, Ningbo, China. (e-mail:Fiseha-Berhanu.Tesema@nottinham.edu.cn)}
\thanks{ Alejandro Guerra Manzanares is with School of Computer Science, UNNC, Ningbo, China (e-mail:alejandro.guerra-manzanares@nottingham.edu.cn)}
\thanks{Tianxiang Cui,  is with School of Computer Science, UNNC, Ningbo, China (e-mail:tianxiang.cui@nottingham.edu.cn)}
\thanks{Qian Zhang, is with School of Computer Science, UNNC, Ningbo, China (e-mail:qian.zhang@nottingham.edu.cn)}
\thanks{Moses Solomon, is with Department of Chemical and Environmental Engineering, and the Nottingham Ningbo China Beacons of Excellence Research and Innovation Institute, UNNC, Ningbo, China.(e-mail:moses.solomon@nottingham.edu.cn)}
\thanks{Sean He, is with School of Computer Science, UNNC, Ningbo, China (e-mail:sean.he@nottingham.edu.cn)}
}

\maketitle

\begin{abstract}

Colorectal cancer (CRC) is a major global cause of cancer-related deaths, with early polyp detection and removal during colonoscopy being crucial for prevention. While deep learning methods have shown promise in polyp segmentation, challenges such as high computational costs, difficulty in segmenting small or low-contrast polyps, and limited generalizability across datasets persist. To address these issues, we propose LGPS, a lightweight GAN-based framework for polyp segmentation. LGPS incorporates three key innovations: (1) a MobileNetV2 backbone enhanced with modified residual blocks and Squeeze-and-Excitation (ResE) modules for efficient feature extraction; (2) Convolutional Conditional Random Fields (ConvCRF) for precise boundary refinement; and (3) a hybrid loss function combining Binary Cross-Entropy, Weighted IoU Loss, and Dice Loss to address class imbalance and enhance segmentation accuracy. LGPS is validated on five benchmark datasets and compared with state-of-the-art(SOTA) methods.  On the largest and challenging PolypGen test dataset, LGPS achieves a Dice of 0.7299 and an IoU of 0.7867, outperformed all SOTA works and  demonstrating robust generalization. With only 1.07 million parameters, LGPS is 17 times smaller than the smallest existing model, making it highly suitable for real-time clinical applications. Its lightweight design and strong performance underscore its potential for improving early CRC diagnosis. Code is available at \href{https://github.com/Falmi/LGPS/}{https://github.com/Falmi/LGPS/}.
\end{abstract}


%
\IEEEpeerreviewmaketitle

\section{Introduction}

Colorectal cancer (CRC) is one of the most prevalent and deadly forms of cancer worldwide, accounting for 10\% of all cancer-related deaths \cite{Yao2021}. Early detection and removal of polyps during colonoscopy are critical to preventing the progression of CRC. However, manually identifying and segmenting polyps in colonoscopy images is challenging due to their significant variability in size, shape, color, and texture, along with the presence of image artifacts such as motion blur, reflectance, and low contrast with surrounding tissues \cite{Kayhan2020, Ali2023}. These challenges often lead to false negatives or inaccurate segmentations, underscoring the need for robust and reliable Computer-Aided Diagnosis (CAD) systems.

In recent years, deep learning-based approaches, particularly Convolutional Neural Networks (CNNs), have shown remarkable success in automating polyp segmentation in colonoscopy images \cite{Fang2019, Fan2020}. Models such as U-Net \cite{Siddique2021}, PraNet \cite{Fan2020}, HarDNet-MSEG \cite{Huang2021}, and WDFF-Net \cite{cao2024wdff} have achieved remarkable segmentation accuracy by leveraging encoder-decoder architectures, attention mechanisms, and multi-scale feature fusion. Despite their effectiveness, these methods face three significant limitations: (i) high computational cost, making them unsuitable for real-time applications \cite{Dong2021}; (ii) difficulty in segmenting small or low-contrast polyps  \cite{Tomar2022}; and (iii) poor generalization across datasets due to variations in imaging conditions \cite{Ali2023}. These limitations hinder the adoption of these segmentation methods in clinical practice.

\begin{figure}[h]
\centering
\includegraphics[width=\columnwidth]{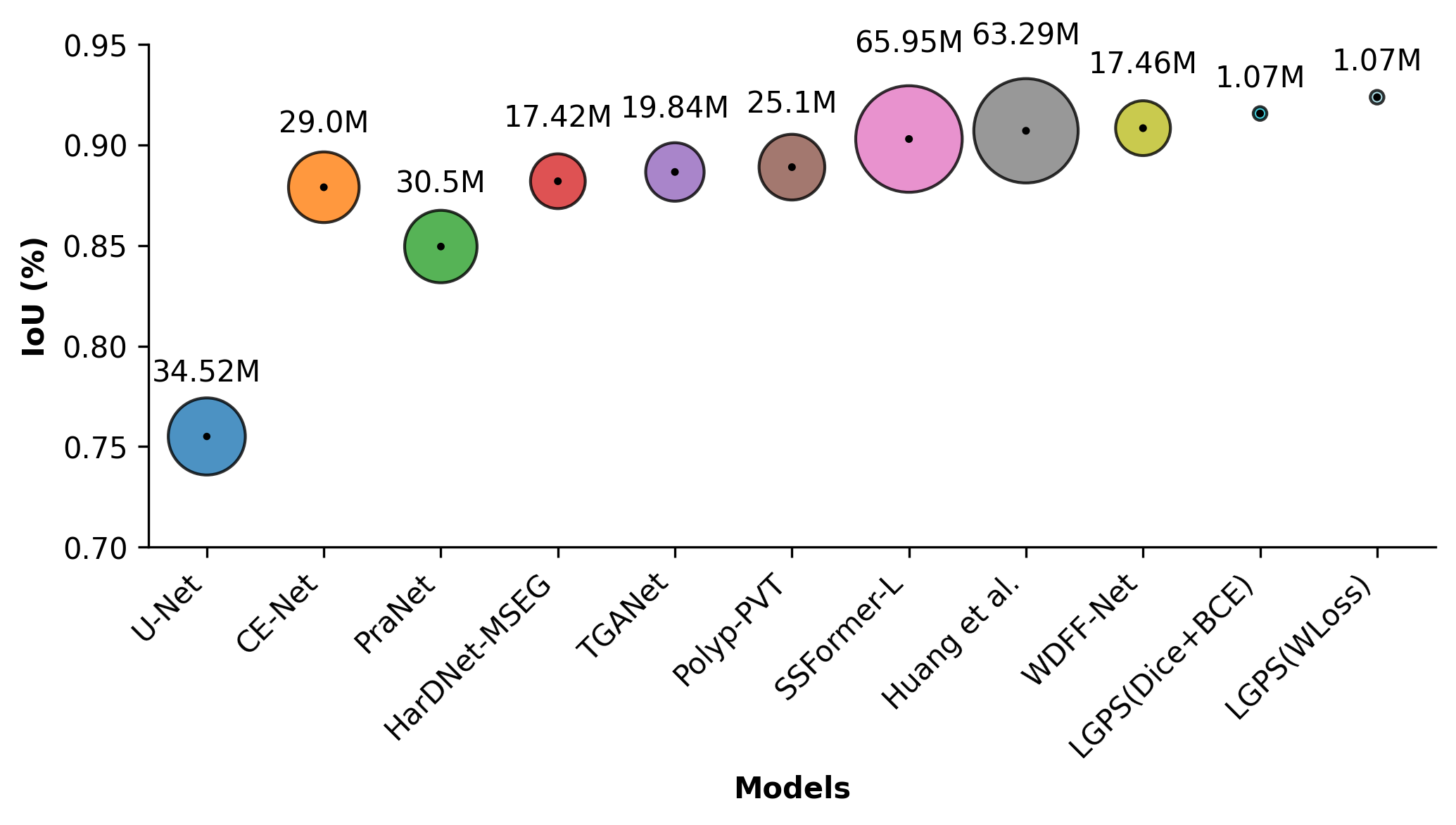}
\caption{\textbf{Comparison of model size and performance.} The area of the circles relates to the size of the model in terms of the number of parameters, while the left axis reports the IoU value of each model on the CVC-ClinicDB dataset. The proposed model, LGPS, outperforms all state-of-the-art models with 17 times fewer parameters.}
\label{fig:Model_parameter}
\end{figure}

To address these challenges, we propose the Lightweight GAN-based framework for Polyp Segmentation (LGPS). As shown in Figure~\ref{fig:Model_parameter}, LGPS achieves state-of-the-art (SOTA) segmentation accuracy with only 1.07 million parameters, making it 17 times smaller than the smallest existing SOTA method. The framework introduces three key innovations: (1) a GAN architecture with a generator for image segmentation and a discriminator enhanced with Convolutional Conditional Random Fields (ConvCRF) to refine spatial coherence and boundary details; (2) a custom hybrid loss function combining Binary Cross-Entropy (BCE), weighted IoU, and Dice losses to address class imbalance and ensure precise segmentation; and (3) a MobileNetV2-based generator with modified residual Squeeze-and-Excitation (ReSE) blocks, enabling SOTA performance while maintaining a lightweight design suitable for deployment on resource-constrained devices. LGPS also demonstrates robust generalization across internal and external validation datasets, including the challenging Polyp-Gen dataset, outperforming larger and more complex models. Its lightweight design and superior accuracy underscore its potential for real-time clinical applications.


\section{Related Work}
\label{sec:related_work}

The rapid development of deep learning methods has significantly advanced computer vision, particularly in medical image segmentation. In this section, we provide an overview of major progress in medical image segmentation methods, focusing on polyp segmentation research. We highlight the strengths and limitations of existing approaches, which emphasize the need for lightweight, efficient, and generalizable models suitable for real-time clinical applications.

\subsection{U-Net Architectures}

The domain of medical image segmentation has made remarkable progress with the advent of deep learning. U-Net \cite{Ronneberger2015}, introduced in 2015, revolutionized the field with its encoder-decoder architecture and skip connections, enabling precise localization and segmentation of medical structures. Since then, U-Net has become a benchmark for many segmentation tasks due to its simplicity and effectiveness. However, it struggles with complex visual patterns, such as polyps with blurry edges or low contrast, often leading to under-segmentation or over-segmentation \cite{vazquez2017benchmark}.

To address these limitations, subsequent works have introduced advanced architectures and techniques. CE-Net \cite{gu2019net} enhances U-Net by incorporating a context encoder module to capture global context information, improving segmentation accuracy in challenging regions. Similarly, PraNet \cite{Fan2020} incorporates parallel reverse attention modules to focus on boundary cues and region relationships, achieving SOTA performance in polyp segmentation tasks. Despite their improvements, these models often require significant computational resources, making them unsuitable for real-time applications or deployment on resource-constrained environments \cite{Huang2021}.

\subsection{Attention Mechanisms and Feature Aggregation}

Attention mechanisms have emerged as a powerful tool for improving segmentation accuracy by enabling models to focus on diagnostically relevant regions \cite{surveymis}. HarDNet-MSEG \cite{Huang2021} uses a cascaded partial decoder and the HarDNet68 \cite{Chao2019} backbone to achieve high accuracy and inference speed, making it suitable for real-time applications. SANet \cite{Fan2017} introduces a shallow attention module to address pixel imbalance in small polyps, effectively reducing background noise and improving segmentation accuracy. TGANet \cite{Tomar2022} leverages text-based embeddings and auxiliary classification tasks to handle drastic scale variations in polyp size. While these methods demonstrate remarkable results, their computational complexity remains a significant drawback \cite{Wang2022}.

\subsection{Transformer-Based Approaches}

The emergence of transformer-based models has further advanced the field of medical image segmentation. Transformers excel at capturing long-range dependencies, making them particularly effective for complex segmentation tasks \cite{lipaper2024}. Polyp-PVT \cite{Dong2021} employs a Pyramid Vision Transformer (PVT) to learn robust feature representations, achieving SOTA performance in polyp segmentation. SSFormer \cite{Wang2022} introduces a progressive local decoder to refine segmentation results, while WDFF-Net \cite{cao2024wdff} combines dual-branch feature fusion with progressive and scale-aware strategies to address under-segmentation and size variation. Despite their effectiveness, transformer-based models are often computationally expensive, with large numbers of parameters and slow inference times, limiting their practicality for real-time applications \cite{tomar2022transresu}.

\subsection{Lightweight Models for Real-Time Deployment}

Given the computational challenges of existing methods, there is a growing demand for lightweight models that can deliver competitive performance while maintaining low memory and computational requirements, especially in resource-constrained environments~\cite{surveycs}. Several works in the literature have proposed lightweight architectures for image segmentation, demonstrating the feasibility of efficient and real-time solutions.

Ni et al. \cite{Ni2020} introduced a bilinear attention network with an adaptive receptive field for the segmentation of surgical instruments. Their approach leverages bilinear attention mechanisms to capture fine-grained details while maintaining computational efficiency. Similarly, Wang et al. \cite{Wang2019} proposed LEDNet, a lightweight encoder-decoder network that uses ResNet50 in the encoder block and an attention pyramidal network in the decoder block. LEDNet achieves real-time semantic segmentation with a significant reduction in computational complexity, making it suitable for resource-constrained environments. Another notable contribution is Squeeze U-Net \cite{Beheshti2020}, which is inspired by the U-Net architecture \cite{Ronneberger2015}. Squeeze U-Net achieves a 12$\times$ reduction in model size compared to traditional U-Net while maintaining efficient performance in terms of multiplication-accumulation (MAC) operations and memory usage. This makes it particularly suitable for deployment on devices with limited computational resources. ERFNet \cite{Romera2017} introduced an efficient residual factorized convolutional network for real-time semantic segmentation. ERFNet achieves a balance between accuracy and speed, making it a strong candidate for real-time applications.

These works highlight the potential of lightweight architectures for real-time image segmentation. However, most existing lightweight models are designed for general-purpose segmentation tasks and have not been specifically optimized for polyp segmentation in colonoscopy images. Polyp segmentation presents unique challenges, such as the need to handle small and irregularly shaped polyps, blurry boundaries, and low contrast with surrounding tissues. These challenges require the development of specialized lightweight models tailored to the particularities of polyp segmentation.

Addressing this significant gap, we propose LGPS, an efficient lightweight GAN-based framework designed specifically for polyp segmentation. LGPS achieves SOTA segmentation accuracy without sacrificing computational efficiency, making it suitable for real-time clinical applications.

\begin{figure}[t]
\centering
\includegraphics[width=\columnwidth]{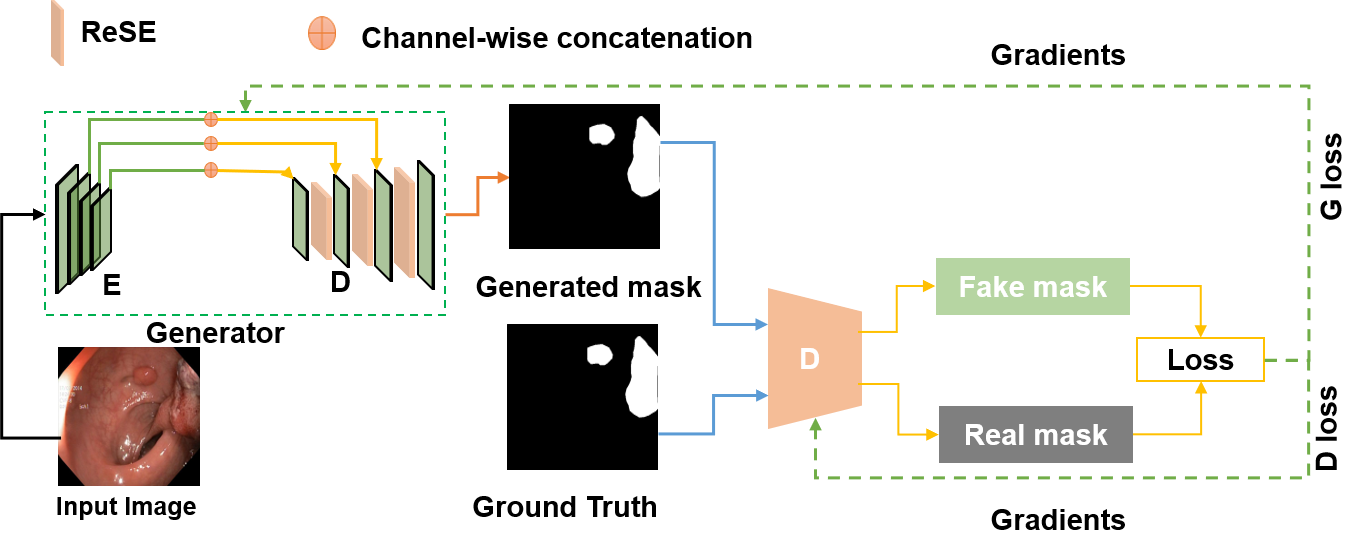}
\caption{a) Overview of the LGPS architecture, showing the generator and discriminator components.}
\label{fig:SegGAN_architecture}
\end{figure}

\section{Proposed Model Architecture}
\label{sec:architecture}

LGPS is a novel GAN-based architecture designed for efficient and precise polyp segmentation in medical images. The model leverages a lightweight backbone, modified Residual Blocks with Squeeze-and-Excitation (ReSE) \cite{hu2018squeeze} mechanisms, and a refinement module to achieve SOTA segmentation accuracy while maintaining computational efficiency. As shown in Figure \ref{fig:SegGAN_architecture}(a), the LGPS architecture consists of two primary components: a generator (G) that produces segmentation masks from input images and a discriminator (D) that evaluates the quality of these masks. In the following, we provide a detailed description of each component, along with theoretical justifications for their design.
\subsection{Generator Architecture}
\label{sec:generator}

The generator follows an encoder-decoder architecture with a modified MobileNetV2 backbone, chosen for its efficiency and lightweight design. The encoder (E) extracts multi-scale features from the input image, while the decoder (D) refines and upsamples these features to produce precise segmentation masks. We perform key modifications to the MobileNetV2~\cite{Sandler2018} backbone to reduce its size to 1.07 million parameters, ensuring computational efficiency for real-time applications. The components of the generator are as follows:

\subsubsection{Encoder}

The encoder is built using a pre-trained MobileNetV2 model, which employs depthwise separable convolutions to reduce computational complexity while maintaining performance. To further optimize the model, we made the following modifications to MobileNetV2: (i) reduce the number of filters in each layer by a factor of 2 to  decrease the number of parameters while retaining essential feature extraction capabilities; (ii) remove the final classification layer and redundant intermediate layers to keep only the essential feature extraction layers for segmentation tasks; and (iii) add depthwise separable convolutions with reduced expansion factors to minimize computational overhead.

These modifications resulted in a lightweight MobileNetV2 backbone with only 1.07 million parameters. The encoder extracts feature maps from four intermediate layers, which are used as skip connections to preserve spatial details during decoding. This multi-scale feature extraction enables the model to handle polyps of varying sizes and shapes, including small polyps that are often missed by other methods.

\begin{figure}[t]
\centering
\includegraphics[height=4cm]{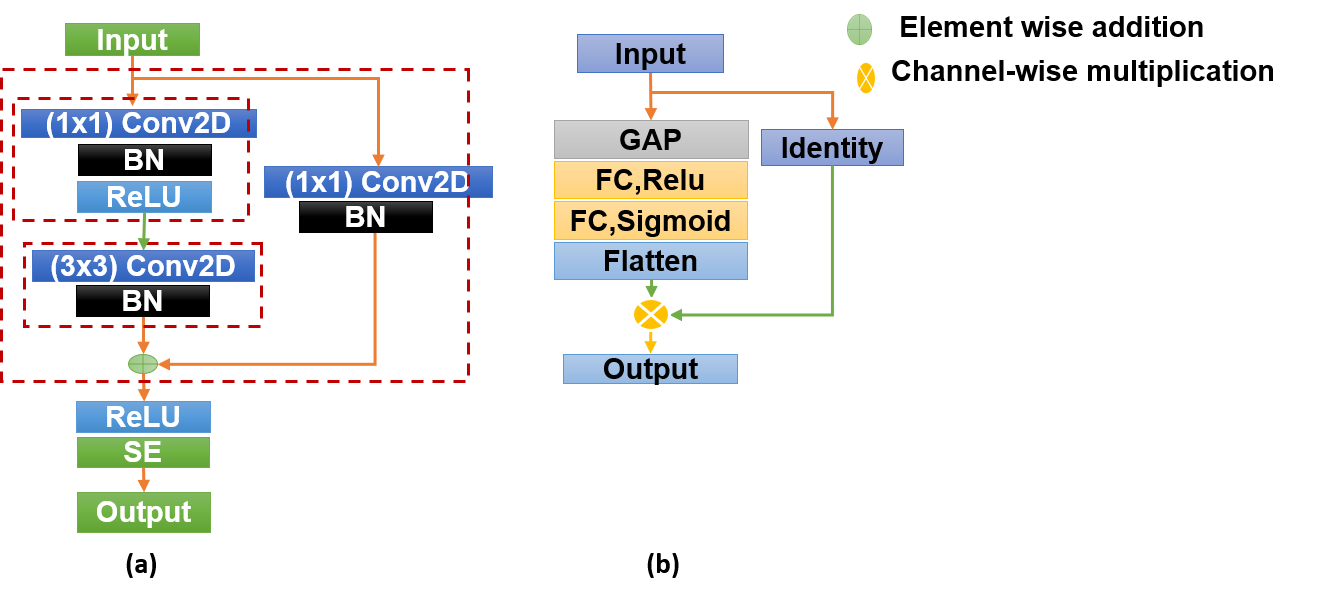}
\caption{a) ReSE block b) SE block}
\label{fig:MRB}
\end{figure}

\subsubsection{Modified Residual with Squeeze-and-Excitation block (ReSE)}
The ReSE block is an essential component of the generator architecture, aiming to enhance feature extraction and recalibration. As shown in Figure~\ref{fig:MRB} (a), it combines traditional residual connections with the SE mechanism, enabling the model to dynamically recalibrate feature maps and focus on diagnostically relevant regions. 

The ReSE block consists of several components that work together to improve feature representation and segmentation performance. They are described as follows.

The first component is the bottleneck layer, which reduces the number of channels by a factor of 4 using a 1x1 convolution layer. This step reduces computational complexity while preserving essential features. The output of the 1x1 convolution is passed through a Batch Normalization layer and a ReLU activation function, expressed as:
\begin{equation}
x_{\text{bottleneck}} = \text{ReLU}(\text{BatchNorm}(\text{Conv2D}_{1\times1}(x))),
\end{equation}
where \( x \) is the input tensor, and \( \text{Conv2D}_{1\times1} \) denotes a 1x1 convolution.

Next, the bottleneck output is passed through a 3x3 convolution layer to extract spatial features. This layer captures local patterns and structures in the feature maps, which are critical for accurate polyp segmentation. The output of the 3x3 convolution is normalized using Batch Normalization:
\begin{equation}
x_{\text{spatial}} = \text{BatchNorm}(\text{Conv2D}_{3\times3}(x_{\text{bottleneck}})).
\end{equation}

A residual connection is then added between the input and the output of the 3x3 convolution to facilitate gradient flow and improve training stability\cite{he2016deep}. If the number of channels in the input tensor does not match the output tensor, a 1x1 convolution is applied to the input tensor to adjust its dimensions:
\begin{equation}
x_{\text{shortcut}} = \text{BatchNorm}(\text{Conv2D}_{1 \times 1}(x)).
\end{equation}
The residual connection is implemented using an Add operation, followed by a ReLU activation:
\begin{equation}
x_{\text{residual}} = \text{ReLU}(\text{Add}([x_{\text{shortcut}}, x_{\text{spatial}}])).
\end{equation}

Finally, the Squeeze-and-Excitation (SE) block recalibrates the feature maps by modeling channel-wise interdependencies\cite{hu2018squeeze}. As show in Figure \ref{fig:MRB} (b), the SE mechanism consists of three steps. First, the squeeze block uses global average pooling to aggregate spatial information into channel-wise descriptors:
\begin{equation}
z_c = \frac{1}{H \times W} \sum_{i=1}^H \sum_{j=1}^W x_c(i, j),
\end{equation}
where \( x_c(i, j) \) is the feature map at spatial location \( (i, j) \) for channel \( c \), and \( H \times W \) is the spatial dimension. Second, the excitation block uses two fully connected (FC) layers to model non-linear channel interdependencies, generating attention weights:
\begin{equation}
s_c = \sigma(W_2 \delta(W_1 z_c)),
\end{equation}
where \( W_1 \) and \( W_2 \) are learnable weights, \( \delta \) is the ReLU activation, and \( \sigma \) is the sigmoid activation. Third, the recalibration block applies the attention weights \( s_c \) to the input feature maps using a Multiply operation, emphasizing diagnostically relevant features:
\begin{equation}
x_{\text{se}} = \text{Multiply}([x_{\text{residual}}, s]).
\end{equation}

This ReSE block, with its combination of bottleneck layers, spatial feature extraction, residual connections, and SE mechanisms, significantly enhances the generator's ability to extract and recalibrate features for accurate polyp segmentation.

\subsubsection{Decoder}
The decoder progressively upsamples the feature maps and concatenates them with skip connections from the encoder to recover spatial details. It consists of four upsampling stages, each followed by the ReSE. The upsampling is performed using bilinear interpolation, and the skip connections are concatenated with the upsampled feature maps to preserve spatial information. The final output of the decoder is a binary segmentation map, obtained by applying a 1x1 convolution with a sigmoid activation:
\begin{equation}
M_{\text{pred}} = \sigma(\text{Conv2D}(1, (1, 1))(F_d^4)),
\end{equation}
where \( F_d^4 \) is the final feature map from the decoder. The use of skip connections and progressive upsampling ensures that the model preserves fine-grained spatial details, a key strength for accurate polyp segmentation.

\subsection{Discriminator Architecture}
\label{sec:discriminator}

The discriminator employs a patch-based adversarial framework with Convolutional Conditional Random Fields (ConvCRF) refinement to improve spatial consistency in polyp segmentation. It processes concatenated pairs of input images and predicted masks \( (I_{\text{in}}, M_{\text{pred}}) \in \mathbb{R}^{256 \times 256 \times 4} \) through five convolutional layers. Each layer uses a kernel size of \( (3, 3) \), stride 2, and LeakyReLU activation (\( \alpha=0.2 \)), progressively increasing the number of filters from 64 to 512. The final layer produces a patch-wise real/fake probability map, providing fine-grained feedback to the generator.

To address spatial inconsistency in GAN-based segmentation, we introduce ConvCRF layers, which refine local spatial coherence through learnable \( 3 \times 3 \) convolutions. A ConvCRF layer consists of a \( 3 \times 3 \) convolutional operation followed by a sigmoid activation:
\begin{equation}
\text{ConvCRF}(F_d^i) = \sigma(\text{Conv}_{3 \times 3}(F_d^i)),
\end{equation}
where \( F_d^i \) represents the feature maps from the previous layer, and \( \sigma \) is the sigmoid activation function. This operation enforces smoothness in predicted masks while preserving edges.

In our implementation, four ConvCRF layers are applied sequentially after the final convolutional layer for refinement. The refined feature maps are computed as:
\begin{equation}
F_{\text{refined}} = \text{ConvCRF}(F_d^i),
\end{equation}
where \( F_d^i \) represents the feature maps from the final convolutional layer. This ensures smoothness and edge preservation in predicted masks.

The final output is a patch-wise real/fake probability map:
\begin{equation}
D(x) = \sigma(F_{\text{refined}}),
\end{equation}
where \( D(x) \) represents the discriminator's output probability. The discriminator is trained to distinguish between real (ground truth) and generated masks, providing adversarial feedback to the generator.

\subsection{Adversarial Training and Loss Functions}
\label{sec:training}

The generator and discriminator are trained in an adversarial manner, where the generator aims to minimize the difference between real and generated masks, while the discriminator attempts to correctly classify real and fake masks. The training process follows a minimax game, defined as:
\begin{equation}
\min_G \max_D \mathcal{L}_{\text{total}}(G, D),
\end{equation}
where \( G \) and \( D \) represent the generator and discriminator, respectively. The generator's loss function is a weighted combination of Binary Crossentropy Loss (BCE), Weighted Intersection over Union (IoU) Loss, and Dice Loss:
\begin{equation}
\mathcal{L}_{\text{total}} = \lambda_1 \mathcal{L}_{\text{BCE}} + \lambda_2 \mathcal{L}_{\text{IoU}} + \lambda_3 \mathcal{L}_{\text{Dice}}.
\end{equation}
These losses guide the generator to produce accurate and realistic segmentation masks. Below, we describe each component of the hybrid loss function in detail.

\subsubsection{Binary Crossentropy Loss (BCE)}
The BCE measures the pixel-wise difference between the predicted mask \( M_{\text{pred}} \) and the ground truth mask \( M_{\text{true}} \). It is defined as:
\begin{multline}
\mathcal{L}_{\text{BCE}} = -\frac{1}{N} \sum_{i=1}^N \Bigl[ M_{\text{true}}^i \log(M_{\text{pred}}^i) \\
+ (1 - M_{\text{true}}^i) \log(1 - M_{\text{pred}}^i) \Bigr],
\end{multline}
where \( N \) is the total number of pixels, and \( M_{\text{true}}^i \) and \( M_{\text{pred}}^i \) are the ground truth and predicted values for the \( i \)-th pixel, respectively.

\subsubsection{Weighted IoU(WIoU) Loss}
The Weighted IoU Loss addresses class imbalance by assigning different weights to the foreground (polyps) and background regions. It is defined as:
\begin{equation}
\mathcal{L}_{\text{IoU}} = 1 - \text{WIoU}(M_{\text{true}}, M_{\text{pred}}),
\end{equation}
where the Weighted IoU is computed as:
\begin{equation}
\text{WIoU}(M_{\text{true}}, M_{\text{pred}}) = \alpha \cdot \text{IoU}_{\text{fg}} + (1 - \alpha) \cdot \text{IoU}_{\text{bg}}.
\end{equation}
Here, \( \alpha \) is the weight for the foreground (typically set to 0.7), and \( \text{IoU}_{\text{fg}} \) and \( \text{IoU}_{\text{bg}} \) are the IoU values for the foreground and background, respectively. These are computed as:
\begin{equation}
\text{IoU}_{\text{fg}} = \frac{\sum (M_{\text{true}} \cdot M_{\text{pred}}) + \epsilon}{\sum M_{\text{true}} + \sum M_{\text{pred}} - \sum (M_{\text{true}} \cdot M_{\text{pred}}) + \epsilon},
\end{equation}

\begin{equation}
\text{IoU}_{\text{bg}} = \frac{A + \epsilon}{B + C - A + \epsilon},
\end{equation}
where \(A = \sum \bigl( (1 - M_{\text{true}}) \cdot (1 - M_{\text{pred}}) \bigr)\), \(B = \sum (1 - M_{\text{true}})\), \(C = \sum (1 - M_{\text{pred}})\), and \(\epsilon\) is a small constant (e.g., \(10^{-6}\)) to avoid division by zero.

The weighted IoU loss ensures that the model focuses on both foreground and background regions, addressing the challenge of class imbalance.

\subsubsection{Dice Loss}
The Dice Loss measures the overlap between the predicted mask and the ground truth mask. It is defined as:
\begin{equation}
\mathcal{L}_{\text{Dice}} = 1 - \frac{2 \sum (M_{\text{true}} \cdot M_{\text{pred}}) + \epsilon}{\sum M_{\text{true}} + \sum M_{\text{pred}} + \epsilon},
\end{equation}
where \( \epsilon \) is a small constant to ensure numerical stability. The Dice Loss is particularly effective for segmentation tasks with imbalanced classes, as it emphasizes the overlap between the predicted and ground truth masks.

\subsubsection{Total Loss}
The total loss for the generator is a weighted combination of the BCE, Weighted IoU, and Dice Losses:
\begin{equation}
\mathcal{L}_{\text{total}} = \lambda_1 \mathcal{L}_{\text{BCE}} + \lambda_2 \mathcal{L}_{\text{IoU}} + \lambda_3 \mathcal{L}_{\text{Dice}}.
\end{equation}
The weights \( \lambda_1 \), \( \lambda_2 \), and \( \lambda_3 \) are hyperparameters that balance the contributions of each loss term. In our experiments, we set \( \lambda_1 = 0.4 \), \( \lambda_2 = 0.3 \), and \( \lambda_3 = 0.3 \) to achieve a balance between pixel-wise accuracy, segmentation overlap quality, and boundary precision.

\subsubsection{Discriminator Loss}
The discriminator is trained using BCE to classify real and fake masks:
\begin{multline}
\mathcal{L}_{D}(y_{\text{true}}, y_{\text{pred}}) = - \frac{1}{N} \sum_{i=1}^{N} \Bigl( y_{\text{true}}^i \log(D(y_{\text{pred}}^i)) \\
+ (1 - y_{\text{true}}^i) \log(1 - D(y_{\text{pred}}^i)) \Bigr),
\end{multline}
where \( y_{\text{true}} \) and \( y_{\text{pred}} \) are the ground truth and predicted labels, respectively, and \( D \) is the discriminator's output probability. The adversarial training framework encourages the generator to produce precise and realistic segmentation masks.

\begin{table}
\centering
\caption{Performance Evaluation Metrics}
\label{tab:Evaluation_Metr}
\begin{tabular}{c c}
\hline
\textbf{Metrics} & \textbf{Description} \\
\hline
Dice & Dice = (2$\times$TP)/(2$\times$TP+FP+FN) \\
IoU & IoU = (TP)/(TP+FP+FN) \\
Recall & Recall = (TP)/(TP+FN) \\
Precision & Precision = (TP)/(TP+FP) \\
Accuracy & Accuracy = (TP+TN)/(TP+TN+FP+FN) \\
F2 & F2 = (5 $\times$ P $\times$R)/(4 $\times$ P + R)\\ \hline
\end{tabular}
\end{table}

\section{ Experimental Result and Analysis}
\label{sec:experiments}

\subsection{Dataset and Evaluation Metrics}
The experiments utilize six public polyp segmentation datasets: Kvasir-SEG \cite{jha2020kvasir}, CVC-ClinicDB \cite{tajbakhsh2015automated}, ETIS \cite{silva2014toward}, CVC-300 \cite{vazquez2017benchmark}, and PolypGen \cite{Ali2023}. These datasets vary in terms of the number of images and their resolutions. Kvasir-SEG contains 1,000 images with variable sizes, while CVC-ClinicDB provides 612 images at a fixed resolution of $384 \times 288$. CVC-ColonDB includes 380 images with a resolution of $574 \times 500$, and ETIS consists of 196 images at a higher resolution of $1225 \times 966$. CVC-300 offers 60 images with the same resolution as CVC-ColonDB ($574 \times 500$), and PolypGen, the largest dataset, contains 1,537 images with variable sizes. These datasets collectively provide a diverse and comprehensive foundation for the experiments.

The performance of the LGPS model was evaluated using a  Dice coefficient (Dice), Intersection over Union (IoU), Recall, Precision, F2 score, and Accuracy. The
formulas for calculating each metric are shown in Table \ref{tab:Evaluation_Metr}.
\subsection{Implementation Details}

The proposed LGPS model is implemented using the TensorFlow and Keras frameworks. The model is trained on a NVIDIA RTX A6000 GPU. The Adam optimizer is used with a learning rate of \(1 \times 10^{-4}\) and a batch size of 16. The Adam optimizer is also used for the discriminator with a learning rate of \(1 \times 10^{-4}\) and a batch size of 16. Input images are preprocessed by resizing them to a fixed resolution of \(256 \times 256\) pixels and normalizing pixel values to the range \([0, 1]\). To improve the robustness of the model, several data augmentation techniques are applied during training. These include random horizontal and vertical flips with a probability of 0.5, random rotation by an angle between \(-10^\circ\) and \(10^\circ\), random brightness adjustment by a factor between 0.9 and 1.1, and random contrast adjustment by a factor between 0.9 and 1.1. The testing set is not augmented and is directly resized into 256 × 256. Following the PraNet \cite{Fan2020} 900 and 550 images from the Kvasir-SEG and CVC-ClinicDB datasets, respectively, are used as the training set, while the remaining 100 and 62 images are used as the testing set. 

\subsection{Ablation Experiments}

\begin{table}
\centering
\caption{Performance of different loss function combinations in the ablation experiments.}
\label{tab:ablation_results}
\begin{tabular}{l c c c c c c }
\hline
Loss Fun. & Dice& IoU& Recall& Pre.& F2 & Acc.\\ \hline
WIoU & 0.8530 & 0.8436 & 0.7900 & 0.9123 & 0.8118 & 0.9498 \\
BCE\_only & 0.8552 & 0.8464 & 0.7866 & 0.9202 & 0.8101 & 0.9506 \\
Dice\_only & 0.8494 & 0.8407 & 0.7722 & 0.9494 & 0.7990 & 0.9494 \\
BWIoU & 0.8515 & 0.8396 & 0.8145 & 0.8816 & 0.8271 & 0.9477 \\
BIoU & 0.8529 & 0.8431 & 0.7954 & 0.9052 & 0.8152 & 0.9494 \\
BDice & \textbf{0.8582} & \textbf{0.8478} & \textbf{0.8049} & 0.9063 & \textbf{0.8233} & 0.9508 \\
3Loss A & 0.8575 & 0.8477 & 0.7905 & 0.9217 & 0.8136 & \textbf{0.9512} \\
3Loss B & 0.8431 & 0.8331 & 0.7880 & 0.8816 & 0.8141 & 0.9457 \\
3Loss C & 0.8377 & 0.8289 & 0.7775 & 0.8952 & 0.7986 & 0.9450 \\ \hline
\end{tabular}
\end{table}
\subsubsection{Ablation Experiment on Loss Function}
To evaluate the impact of different loss functions, we conducted an ablation study using the Kvasir-SEG dataset. We tested various combinations of Binary Cross-Entropy (BCE), Intersection over Union (IoU), Weighted IoU (WIoU), and Dice Loss, as summarized in Table~\ref{tab:ablation_results}. The results reveal that the combination of BCE and Dice Loss (BDice) achieved the highest Dice score (0.8582) and IoU (0.8478), outperforming other combinations. Below, we discuss the performance of standalone and combined loss functions.

Standalone loss functions address one specific aspect of the segmentation task. Weighted IoU (WIoU) achieved a Dice score of 0.8530 and IoU of 0.8436. WIoU balances foreground and background regions, addressing class imbalance but lacks pixel-wise accuracy and boundary precision. BCE Only achieved a Dice score of 0.8552 and IoU of 0.8464. While BCE Loss ensures pixel-wise classification accuracy, it struggles with class imbalance and boundary precision. Dice Only achieved a Dice score of 0.8494 and IoU of 0.8407. Dice Loss handles class imbalance and optimizes overlap but lacks pixel-wise precision.

Combined loss functions address multiple aspects of the segmentation task by integrating two or more losses. The BCE + Dice Loss (BDice) combination achieved the highest Dice score (0.8582) and IoU (0.8478). This combination balances pixel-wise accuracy (BCE) with overlap quality and boundary precision (Dice), addressing class imbalance and producing well-defined boundaries. BCE + WIoU (BWIoU) achieved a Dice score of 0.8515 and IoU of 0.8396. While BWIoU improves over standalone WIoU by balancing pixel-wise accuracy and foreground-background balancing, it does not explicitly optimize for boundary precision. BCE + IoU (BIoU) achieved a Dice score of 0.8529 and IoU of 0.8431. This combination balances pixel-wise accuracy with overlap metrics but lacks the boundary refinement provided by Dice Loss.

Hybrid loss combinations, such as 3Loss A (\(0.4 \cdot \text{BCE} + 0.3 \cdot \text{WIoU} + 0.3 \cdot \text{Dice}\)), achieved competitive results, with a Dice score of 0.8575 and IoU of 0.8477. While this hybrid loss balances pixel-wise accuracy, foreground-background balancing, and overlap quality, it is slightly outperformed by BDice, suggesting that the additional complexity of combining three losses does not always translate to better performance. Similarly, 3Loss B (\(0.3 \cdot \text{BCE} + 0.4 \cdot \text{WIoU} + 0.3 \cdot \text{Dice}\)) achieved a Dice score of 0.8431 and IoU of 0.8331. This combination places more emphasis on WIoU, reducing its effectiveness in handling boundary precision and pixel-wise accuracy. 3Loss C (\(\text{BCE} + \text{IoU} + \text{Dice}\)) achieved a Dice score of 0.8377 and IoU of 0.8289. This combination lacks the weighted balancing of foreground and background regions, which reduces its effectiveness in handling class imbalance.

In conclusion, the BCE + Dice Loss (BDice) combination is the most effective for polyp segmentation, as it addresses the key challenges of class imbalance, boundary precision, and pixel-wise accuracy without introducing unnecessary complexity. Standalone losses and hybrid combinations, while useful in specific scenarios, do not outperform the simpler BCE + Dice combination.

\begin{table}
\centering
\caption{Ablation experiment on the contribution of each modules.} 
\begin{tabular}{c c c}
\hline
Model & Dice& IoU \\ \hline
Baseline & 0.8575 & 0.8477 \\ 
W/o ReSE & 0.8445 & 0.8366 \\ 
W/o ConvCRF& 0.8519 & 0.8436 \\ 
MRB w/o SE & 0.8475 & 0.8378 \\ 
W/o ConvCRF and ReSE & 0.8415 & 0.8376 \\ \hline
\end{tabular}

\label{tab:ablations}
\end{table}

\begin{figure}[h!]
\centering
\includegraphics[height=3cm]{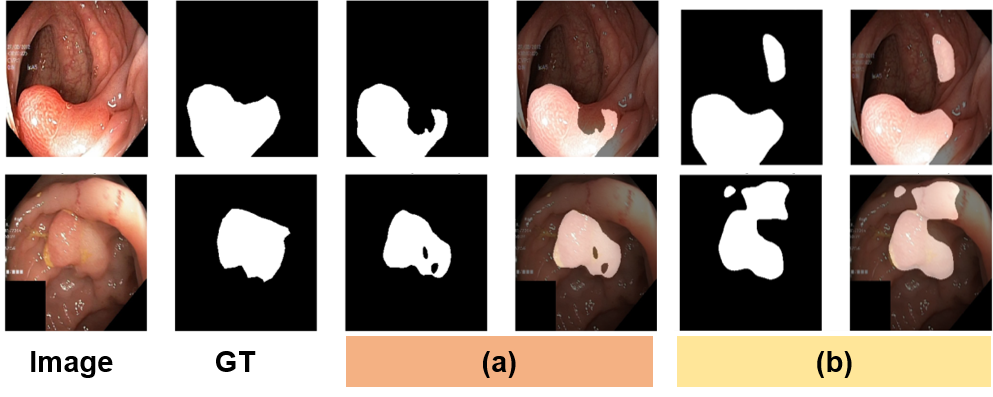}
\caption{Visualized heat maps (a) with ConvCRF and ReSE and (b) without ConvCRF and ReSE}
\label{fig:heat_map_with_without}
\end{figure}

\begin{figure*}[h!]
\centering
\includegraphics[height=3cm]{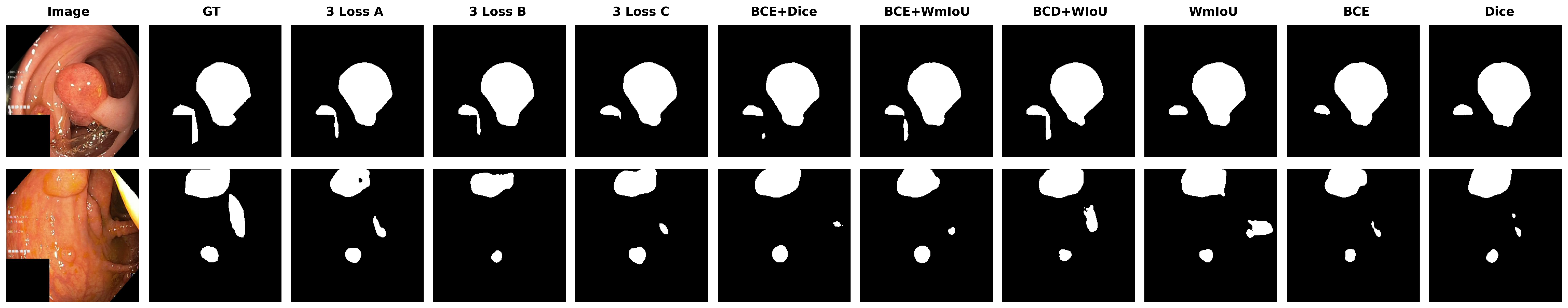}
\caption{Ablation experiment on different loss functions.}
\label{fig:Loss_Abilation}
\end{figure*}

\subsubsection{Ablation Experiment on the Contribution of Each Module}

To evaluate the contribution of each component in LGPS, we conducted an ablation study using the 3Loss A as the benchmark, we named it LGPS Weighted Loss (LGPS WLoss). The study systematically removed key components and analyzed their impact on segmentation performance. The results, presented in Table~\ref{tab:ablations}, are evaluated using the Dice and  IoU.

The baseline model, which includes all components 
(ReSE, ConvCRF layers, and the WLoss function), achieved the highest performance with a Dice of 0.8575 and IoU of 0.8477. This result demonstrates the effectiveness of the complete model configuration, where each component contributes to improving segmentation accuracy and spatial coherence.

When the ReSE was removed, the Dice dropped to 0.8445 (a reduction of 1.30\%), and the IoU decreased to 0.8366 (a reduction of 1.11\%). This performance degradation highlights the importance of the MRB in capturing hierarchical features and enhancing the model's ability to handle complex polyp structures. The residual connections within these blocks facilitate gradient flow during training, enabling the model to learn more robust representations.

Removing the SE mechanism from the ReSE resulted in a Dice of 0.8475 (a reduction of 1.00\%) and an IoU of 0.8378 (a reduction of 0.99\%). The SE mechanism dynamically recalibrates channel-wise feature responses, emphasizing diagnostically relevant features while suppressing less useful ones. Its removal leads to a noticeable drop in performance, particularly in scenarios where fine-grained feature discrimination is critical.

When the ConvCRF layer was removed, the Dice decreased to 0.8519 (a reduction of 0.56\%), and the IoU dropped to 0.8436 (a reduction of 0.41\%). The ConvCRF layer plays a crucial role in refining segmentation masks by enforcing spatial coherence and preserving boundary details. Its removal results in slightly less precise segmentation, particularly around polyp edges and small structures.

Removing both the ConvCRF layer and the ReSE led to the most significant performance degradation, with the Dice dropping to 0.8415 (a reduction of 1.60\%) and the IoU decreasing to 0.8376 (a reduction of 1.01\%). This result underscores the complementary roles of these components: the ReSE block enhances feature extraction, while the ConvCRF layer refines the final segmentation output. Their combined removal significantly impacts the model's ability to accurately segment polyps, even with the WLoss function in place.

The ablation study demonstrates that each component of LGPS contributes meaningfully to its overall performance. The ReSE mechanism are critical for robust feature extraction, while the ConvCRF layer ensures precise boundary preservation and spatial coherence. The baseline model, which includes all components, achieves the best performance, highlighting the importance of their synergistic integration.

\subsubsection{Qualitative Ablation Study: Impact of Key Components on Segmentation Performance}
The heat map of the features, both with and without the ConvCRF and ReSE, is shown in \ref{fig:heat_map_with_without}. It is evident that the network focuses more on the object areas when both modules are introduced. Without these modules, the network activates non-polyp regions and fails to precisely localize the polyp region and its shape. However, with the inclusion of both modules, the polyp regions are accurately activated. This indicates that the modules enhance the object regions while suppressing the background, thereby improving segmentation accuracy.

\subsubsection{Qualitative Analysis of Segmentation Masks: Evaluating the Impact of Different Loss Functions}
\label{sec:qualitative}

To complement the quantitative findings of the ablation study, as shown in Figure~\ref{fig:Loss_Abilation}, we performed a qualitative analysis of the segmentation masks generated by the different loss functions. This analysis focused on visual inspection of the segmentation results, particularly for challenging cases such as small polyps, boundary regions, and areas with class imbalance.

The qualitative analysis revealed several key observations. The segmentation masks produced by the combination 3Loss A exhibited the highest precision in boundary localization. The edges of the polyps were well-defined, and the masks closely aligned with the ground truth annotations, even in regions with complex shapes or irregular boundaries. For small polyp regions, the baseline 3Loss A and 3Loss B demonstrated superior performance. The segmentation masks generated by these loss functions accurately captured small polyps, with minimal false positives or missed regions. This aligns with the quantitative results from the ablation study, confirming their effectiveness in handling small and underrepresented structures.

The WLoss functions, particularly the baseline and BCD + WIoU, showed a remarkable ability to handle class imbalance. In images with a high background-to-polyp ratio, these loss functions produced segmentation masks that effectively prioritized polyp regions without over-segmenting the background. However, standalone loss functions such as WIoU, BCE, and Dice loss, while effective in segmenting large polyp regions, exhibited limitations in generalizing across diverse cases. For instance, they occasionally produced fragmented masks for small polyps or failed to predict small polyp regions altogether. Additionally, these standalone loss functions struggled with boundary precision in regions of low contrast.

Some loss functions, particularly standalone and binary loss functions, exhibited tendencies toward over-segmentation or under-segmentation. Over-segmentation was observed in regions with ambiguous boundaries, while under-segmentation occurred in cases where the polyp regions were small or poorly contrasted against the background. These challenges highlight the limitations of using single loss functions in complex segmentation tasks.

The qualitative analysis underscores the strengths and limitations of the evaluated loss functions in polyp segmentation. The combination 3Loss A consistently demonstrated superior performance in boundary precision, small polyp localization, and handling class imbalance. Standalone loss functions, while effective for large polyps, struggled with small polyp regions and boundary precision. These findings reinforce the importance of combining multiple loss functions to address the diverse challenges in polyp segmentation and provide valuable insights for future improvements in segmentation models.

\begin{table*}[ht]
\centering
\begin{tabular}{l l c c c c c c}
\hline
Method & Backbone Network & Parameters (M) & Dice & IoU & Recall & Precision & F2 \\ \hline
U-Net \cite{Zhuang2019} & ResNet34 & 34.52 & 0.8762 & 0.7550 & 0.8732 & 0.8999 & 0.8784 \\
CE-Net \cite{gu2019net} & ResNet34 & 29.00 & 0.9280 & 0.8790 & 0.9080 & 0.9150 & 0.8990 \\
PraNet \cite{Fan2020} & Res2Net & 30.50 & 0.8995 & 0.8495 & 0.9500 & 0.9450 & 0.9490 \\
HarDNet-MSEG \cite{huang2021hardnet} & HardNet68 & 17.42 &0.9320 & 0.8820 & 0.9200 & 0.9460 & 0.9290\\
TGANet \cite{Tomar2022} & ResNet50 & 19.84 & 0.9457 & 0.8866 & 0.9437 & 0.9519 & 0.9439 \\
Polyp-PVT \cite{Dong2021} & PVT & 25.10 & 0.9370 & 0.8890 & 0.9490 & 0.9280 & 0.9360 \\
SSFormer-L \cite{Wang2022} & PVT & 65.95 & 0.9470 & 0.9030 & 0.9560 & 0.9420 & 0.9530 \\
Huang et al. \cite{Huang2022} & ResNet50 & 63.29 & 0.9492 & 0.9071 & 0.9534 & 0.9483 & 0.9511 \\
WDFF-Net \cite{cao2024wdff} & HardNet68 & 17.46 & \textbf{0.9521} & 0.9084 & \textbf{0.9702} & \textbf{0.9711} & \textbf{0.969} \\ 
\textbf{Ours (Weighted Loss)} & MobileNet-V2 & 1.07 & 0.9261 & \textbf{0.9238} & 0.8607 & 0.9683 & 0.8802 \\ 
\textbf{Ours (Dice + BCE)} & MobileNet-V2 & 1.07 & 0.9117 & 0.9157 & 0.8473 & 0.9655 & 0.8686 \\ \hline
\end{tabular}
\caption{Comparison with SOTA methods on the CVC-ClinicDB dataset.}
\label{tab:CVCC}
\end{table*}

\begin{table}[ht]
\centering
\caption{Comparison with SOTA on ETIS, and CVC-300 datasets.}

\begin{tabular}{c c c| c c}
\hline
\multirow{2}{*}{Method}  & \multicolumn{2}{c|}{ETIS} & \multicolumn{2}{c}{CVC-300} \\ \cline{2-5} 
& \textbf{Dice}  & \textbf{IoU} & \textbf{Dice} & \textbf{IoU} \\ \hline
U-Net \cite{Zhuang2019}  & 0.3980 & 0.3350 & 0.7100 & 0.6270 \\ 
CE-Net \cite{gu2019net}  & 0.5859 & 0.5700 & 0.8706 & 0.7970 \\ 
PraNet \cite{Fan2020} & 0.6280 & 0.5670 & 0.8710 & 0.7970 \\ 
HarDNet-MSEG \cite{huang2021hardnet} & 0.6770 & 0.6630 & 0.8870 & 0.8210 \\ 
TGANet \cite{Tomar2022} & 0.6630 & 0.5860 & 0.8850 & 0.8190 \\ 
Polyp-PVT \cite{Dong2021} & 0.7870 & 0.7600 & 0.9000 & 0.8330 \\ 
Huang et al. \cite{Huang2022} & 0.7510 & 0.6800 & 0.9110 & 0.8490 \\ 
Su et al. \cite{su2023accurate} & \textbf{0.8160} & 0.7330 & 0.9120 & 0.8490 \\ 
WDFF-Net(2024) \cite{cao2024wdff}  & 0.7581 & 0.7241 & \textbf{0.9161} & 0.8533 \\ 
\textbf{Ours (Weighted Loss)} &0.7447 & 0.7742 & 0.8502 & 0.8648\\ 
\textbf{Ours (Dice + BCE)} &0.7451 & \textbf{0.7746} & 0.8556 & \textbf{0.8690} \\ \hline
\end{tabular}
\label{tab:CVC,ETIS}
\end{table}

\begin{table}[ht]
\centering
\caption{Comparison with SOTA methods on the PolypGen dataset.}
\setlength{\tabcolsep}{3pt} 
\begin{tabular}{l l c c c c}
\hline
Method &Dice & IoU & Recall & Pre. & F2 \\ \hline
U-Net (2015) \cite{Zhuang2019} & 0.5995 & 0.5347 & 0.6829 & 0.7523 & 0.6105 \\
U-Net++(2018) \cite{zhou2018unet++} & 0.5964 & 0.5310 & 0.6765 & 0.7546 & 0.6089 \\
ResU-Net++(2019) \cite{jha2019resunet++} & 0.3982 & 0.3149 & 0.5887 & 0.4444 & 0.4314 \\
HarDNet-MSEG(2021) \cite{huang2021hardnet} &0.6089 & 0.5376 &0.7116 & 0.7124 & 0.6246\\
ColonSegNet (2021)\cite{Fan2020} & 0.5486 & 0.4718 & 0.6554 & 0.6687 & 0.5617 \\
UACANet(2021) \cite{kim2021uacanet} & 0.6531 & 0.5777 & 0.7493 & 0.7531 & 0.6678 \\
UNeXt (2022) \cite{valanarasu2022unext} & 0.4552 & 0.3761 & 0.6135 & 0.5600 & 0.4805 \\
TransNetR (2023)\cite{jha2024transnetr} & 0.6668 & 0.6058 & 0.6135 & 0.5600 & 0.6706 \\
WDFF-Net (2024) \cite{cao2024wdff} & 0.6687 & 0.6102 & 0.6893 & 0.7602 & 0.6723 \\ 
\textbf{Ours (WLoss)} & \textbf{0.7299} &\textbf {0.7867} & 0.6807 & \textbf{0.8233} & 0.6958 \\
\textbf{Ours (Dice+BCE)} & 0.7276 & 0.7835 & \textbf{0.6997} & 0.7948 & \textbf{0.7061} \\ \hline
\end{tabular}

\label{tab:comparison_polypgen}
\end{table}

\subsection{Qualitative Assessment of LGPS and Existing Methods}
\label{sec:visualization}
\begin{figure*}[h!]
\centering
\includegraphics[height=7.5cm]{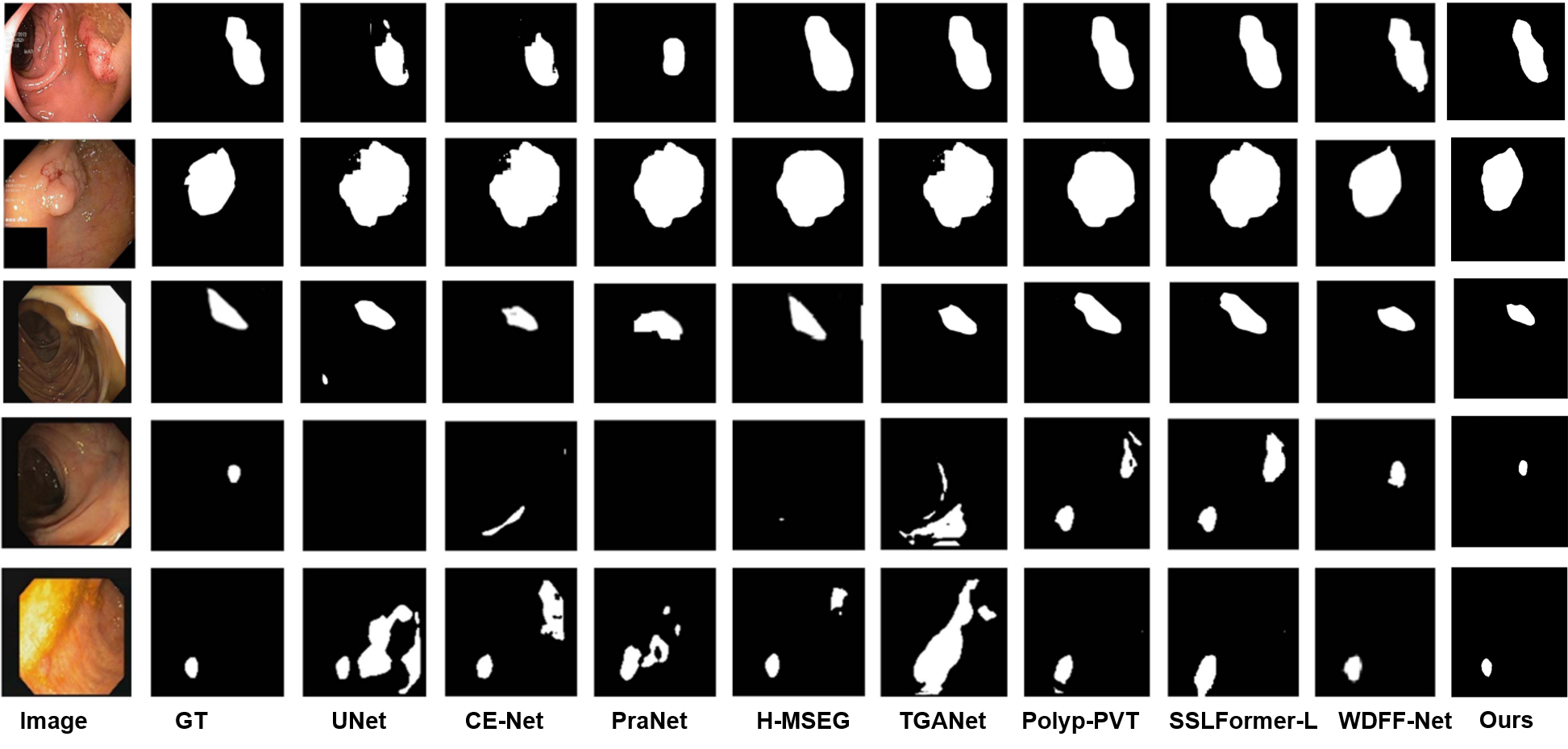}
\caption{Visualization of segmentation results on the Kvasir-SEG and CVC-ColonDB datasets. Rows 1--2 shows large polyps. Rows 3--4 show cases with blurry polyp boundaries and low contrast, while rows 4--5 depict cases with significant variations in polyp size, particularly small polyps.}
\label{fig:vis_resu_different_method}
\end{figure*}

To qualitatively evaluate the performance of different state-of-the-art (SOTA) segmentation methods, we visualize the segmentation results on the Kvasir-SEG and CVC-ColonDB datasets, as shown in Fig.~\ref{fig:vis_resu_different_method}. The visualization highlights the strengths and limitations of existing methods compared to the proposed LGPS.

Existing methods often struggle with challenges such as blurry boundaries and low contrast. As shown in rows 1-2 existing methods over segmented polyp region, however, LGSP precisely segment the polyp region.  As shown in rows 3--4 of Fig.~\ref{fig:vis_resu_different_method}, existing methods frequently fail to achieve complete segmentation when polyp boundaries are blurry or exhibit low contrast with surrounding tissues. This results in incomplete or inaccurate segmentation masks. Additionally, existing methods face difficulties with variable polyp sizes. Rows 4--5 illustrate that existing methods tend to miss small polyps or produce fragmented segmentation results when there are significant variations in polyp size. This is particularly problematic for small polyps, which are often overlooked or inaccurately segmented. In contrast, the proposed LGPS demonstrates superior performance in addressing these challenges. LGPS effectively segments flat polyps with low contrast, as shown in rows 3--4. The model's ability to capture subtle boundary details ensures complete and accurate segmentation, even in challenging cases. Furthermore, LGPS accurately segments polyps with large variations in size, including small polyps, as depicted in rows 4--5. This robustness is attributed to the model's multi-scale feature extraction and boundary refinement mechanisms. However, in the fourth row, LGPS struggles slightly to precisely segment the polyp region under conditions of poor visibility. The visualization experiment demonstrates that LGPS outperforms existing methods in accurately segmenting polyps with blurry boundaries, low contrast, and significant size variations. These results underscore the model's ability to handle real-world challenges in polyp segmentation, making it a reliable tool for clinical applications.

\subsection{State-of-the-Art (SOTA) Analysis and Discussion}

To validate the effectiveness of LGPS, we compared it against nine SOTA polyp segmentation methods, including both universal segmentation networks (e.g., U-Net \cite{Zhuang2019}, CE-Net \cite{gu2019net}) and dedicated polyp segmentation networks (e.g., PraNet \cite{Fan2020}, HarDNet-MSEG \cite{huang2021hardnet}, TGANet \cite{Tomar2022}, SSFormer-L \cite{Wang2022}, Polyp-PVT \cite{Dong2021}, and WDFF-Net \cite{cao2024wdff}). The comparison was conducted on four public datasets: CVC-ClinicDB, ETIS, CVC-300, and PolypGen.

\subsubsection{Segmentation Accuracy}
As shown in Table \ref{tab:CVCC}, LGPS demonstrates competitive segmentation accuracy on the CVC-ClinicDB dataset. The WLoss variant achieves a Dice score of 0.9261 and an IoU of 0.9238, outperforming PraNet (Dice = 0.8995) and achieving results comparable to several SOTA methods. Notably, LGPS achieves the highest IoU (0.9238) among all methods, surpassing even the recent WDFF-Net (IoU = 0.9084). This highlights the effectiveness of LGPS in achieving high segmentation accuracy, particularly when leveraging the WLoss function, which balances multiple loss terms to improve performance.

\subsubsection{Generalization Ability}
One of the key strengths of LGPS is its exceptional generalization capability, particularly on unseen and challenging datasets. To evaluate this, we conducted experiments on three unseen datasets: PolypGen, ETIS, and CVC-300. As shown in Table \ref{tab:CVC,ETIS} and \ref{tab:comparison_polypgen}, LGPS achieves strong performance on ETIS (IoU = 0.7746) and CVC-300 (IoU = 0.8690), demonstrating its robustness to diverse imaging conditions and unseen data. In terms of Dice, LGPS shows competitive results on both datasets, with 0.7447 on ETIS and 0.8502 on CVC-300.

Notably, LGPS achieves SOTA performance on the PolypGen dataset, the largest and most challenging test set, with a Dice score of 0.7299 and an IoU of 0.7867. This is a significant achievement, as PolypGen contains diverse polyp types and imaging conditions, making it a rigorous benchmark for evaluating generalization. To the best of our knowledge, LGPS is the first polyp segmentation model to demonstrate such strong generalization performance on unseen datasets. The adversarial training framework, combined with the WLoss function, enables the model to learn robust features that generalize well across different datasets and imaging conditions.

\subsubsection{Model Efficiency}
A key advantage of LGPS is its lightweight design, enabled by the MobileNet-V2 backbone. With only 1.07 million parameters, LGPS is significantly more efficient than SOTA methods such as WDFF-Net (17.46M parameters), SSFormer-L (65.95M parameters), and Huang et al. (63.29M parameters). Despite its compact architecture, LGPS achieves competitive or superior performance on multiple datasets, making it suitable for real-time applications in clinical settings. This efficiency is particularly important for deploying the model in resource-constrained environments, such as endoscopy suites, where computational resources are limited.

\subsubsection{Discussion}
The results demonstrate that LGPS achieves a compelling balance of accuracy, efficiency, and generalization. The WLoss variant, which combines BCE, WIoU, and Dice Loss, consistently outperforms the Dice+BCE variant, particularly on unseen datasets. This highlights the importance of balancing multiple loss terms to improve segmentation performance and generalization.

The strong performance of LGPS on unseen datasets, particularly PolypGen, can be attributed to its GAN-based architecture. The adversarial training framework encourages the generator to produce realistic and accurate segmentation masks, while the discriminator provides fine-grained feedback to improve boundary preservation and spatial coherence. This makes LGPS particularly effective in handling diverse and challenging imaging conditions, which are common in real-world clinical settings.

While LGPS achieves SOTA performance on PolypGen and SOTA IoU on other datasets, there is room for improvement in terms of Dice, where it slightly underperforms compared to some methods. Future work could explore integrating additional attention mechanisms or leveraging transformer-based backbones to further enhance performance. These improvements could address the current limitations and extend the applicability of LGPS to a wider range of medical imaging tasks.

\section{Conclusion}
\label{sec:conclusion}

In this paper, we introduced LGPS, a lightweight GAN-based framework for polyp segmentation in colonoscopy images. LGPS addresses critical challenges such as blurry boundaries, small polyp detection, and computational inefficiency, making it suitable for real-time clinical applications. The framework integrates a MobileNetV2 backbone with ReSE, and ConvCRF to achieve SOTA performance with only 1.07 million parameters. A hybrid loss function combining Binary Cross-Entropy (BCE), Weighted IoU Loss, and Dice Loss further enhances segmentation accuracy by addressing class imbalance.

Extensive experiments on five public datasets demonstrate the effectiveness of LGPS. On the challenging PolypGen dataset, LGPS achieves a Dice of 0.7299 and a mean IoU of 0.7867, outperforming existing methods in both accuracy and efficiency. The model also exhibits strong generalization capabilities on unseen datasets, such as ETIS and CVC-300 , highlighting its robustness to diverse imaging conditions. Its lightweight design makes it highly suitable for deployment on resource-constrained devices, offering significant potential for real-time clinical use.

Future work will explore extending LGPS to other medical imaging tasks, such as lesion detection and organ segmentation, and integrating transformer-based architectures to further enhance performance. By addressing key challenges in polyp segmentation, LGPS sets a new benchmark for efficient and accurate medical image analysis, paving the way for improved clinical outcomes.

\end{document}